\documentclass[letterpaper]{article} 
\usepackage{aaai25}  
\usepackage{times}  
\usepackage{helvet}  
\usepackage{courier}  
\usepackage[hyphens]{url}  
\usepackage{graphicx} 
\usepackage{natbib}  
\usepackage{caption} 
\usepackage{algorithm}
\usepackage{algorithmic}
\usepackage{amsmath} 

\usepackage{newfloat}
\usepackage{listings}
\DeclareCaptionStyle{ruled}{labelfont=normalfont,labelsep=colon,strut=off} 
\floatstyle{ruled}
\newfloat{listing}{tb}{lst}{}
\floatname{listing}{Listing}
\usepackage[dvipsnames]{xcolor}
\usepackage{array,booktabs,multirow}
\usepackage{caption}
\usepackage{subcaption}

\newcommand\eat[1]{{}}

\newcommand{\FIDLAr}{{\sc FIDLAr}}

\title{\FIDLAr: Forecast-Informed Deep Learning Architecture for Flood Mitigation}
\author{
    Jimeng Shi\textsuperscript{\rm 1},
    Zeda Yin\textsuperscript{\rm 2},
    Arturo Leon\textsuperscript{\rm 2},
    Jayantha Obeysekera\textsuperscript{\rm 2},
    Giri Narasimhan\textsuperscript{\rm 1}
}
\affiliations{
    \textsuperscript{\rm 1}Knight Foundation School of Computing and Information Sciences, Florida International University\\
    \textsuperscript{\rm 2}Department of Civil and Environmental Engineering, Florida International University\\
    \textsuperscript{\rm 3}Sea Level Solutions Center, Florida International University\\
    \{jshi008, zyin005, arleon, jobeysek, giri\}@fiu.edu
}

\begin{document}

\maketitle

\begin{abstract}
In coastal river systems, floods, often during major storms or king tides, severely threaten lives and property. 
However, hydraulic structures such as dams, gates, pumps, and reservoirs exist in these river systems, and these floods can be mitigated or even prevented by strategically releasing water before extreme weather events.
A standard approach used by local water management agencies is the ``rule-based'' method, which specifies predetermined water prereleases based on historical human experience, but which tends to result in excessive or inadequate water release. 
Iterative optimization methods that rely on detailed physics-based models for prediction are an alternative approach. Whereas, such methods tend to be computationally intensive, requiring hours or even days to solve the problem optimally.
In this paper, we propose a \underline{\textbf{F}}orecast \underline{\textbf{I}}nformed \underline{\textbf{D}}eep \underline{\textbf{L}}earning \underline{\textbf{Ar}}chitecture, \FIDLAr, to achieve rapid and near-optimal flood management with precise water prereleases. 
\FIDLAr\ seamlessly integrates two neural network modules: one called the \texttt{Flood Manager}, which is responsible for generating water pre-release schedules, and another called the \texttt{Flood Evaluator}, which evaluates those generated schedules.
The Evaluator module is pre-trained separately, and its gradient-based feedback is utilized to train the Manager model, ensuring near-optimal water pre-releases.
We have conducted experiments with a flood-prone coastal area in South Florida.
Results show that \FIDLAr\ is several orders of magnitude faster than currently used physics-based approaches while outperforming baseline methods with improved water pre-release schedules. 
\begin{links}
\link{Code}{https://github.com/JimengShi/FIDLAR}
\end{links}
\end{abstract}

\section{Introduction}
\label{sec:intro}
%
Floods can result in catastrophic consequences with considerable loss of life \cite{jonkman2008loss}, huge socio-economic impact \cite{wu2021new}, property damage \cite{brody2007rising}, and environmental devastation \cite{yin2023flash}. 
They pose a threat to food and water security as well as sustainable development \cite{kabir2019impacts}. 
What is even more alarming is that research indicates global climate change may lead to a drastic increase in flood risks in terms of both frequency and scale
\cite{wing2022inequitable}, e.g., coastal flood risks due to sea-level rise \cite{sadler2020exploring}. 
Thus, effective flood management is of utmost importance.

To mitigate flood risks, water management agencies have built controllable hydraulic structures such as dams, gates, pumps, and reservoirs in river systems \cite{kerkez2016smarter}. 
However, determining the optimal \textit{control schedules} of these hydraulic structures is a challenging problem \cite{bowes2021flood}. 
Rule-based methods \cite{sadler2019leveraging} formulate control schedules based on insights gained from historically observed data. 
The rules represent the collective wisdom gathered over decades of experience in managing specific river systems.
Nevertheless, these rules may expose vulnerabilities while dealing with extremely rare events and may not offer effective solutions for complex river systems under such conditions \cite{schwanenberg2015open}. 
\begin{figure*}[ht]
\centering
    \includegraphics[width=1.95\columnwidth]{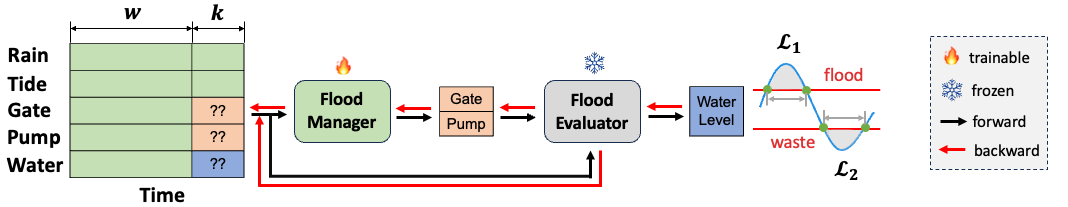} 
\caption{Forecast-Informed Deep Learning Architecture (\FIDLAr). 
Input data consists of five categories of variables as shown in the left table. The variables $w$ and $k$ are the lengths of the past and prediction windows, respectively.
The parts colored green are provided as inputs, while the orange and blue parts (with question marks) are outputs. 
The \texttt{Flood Manager} and \texttt{Flood Evaluator} represent deep learning (DL) models, the former to predict control schedules of controllable hydraulic structures (e.g., gates and pumps) to pre-release water, and the latter to predict the resulting water levels for those control schedules. 
Loss functions, $\mathcal{L}_{1}$ and $\mathcal{L}_{2}$, penalize the \emph{flooding} and \emph{water wastage} beyond pre-specified thresholds, respectively.}
\label{fig:fidla_framework}
\end{figure*}

The flood mitigation problem is usually treated as an optimization task \cite{karimanzira2016model}, where variables such as control schedules for hydraulic structures are optimized to achieve the desired water levels to effectively mitigate flood risks \cite{zarei2021machine}.
Random initialization followed by soft-computing techniques such as genetic algorithm (GA) \cite{leon2020matlab} is often employed to perform the optimizations.
Subsequently, physics-based simulator models such as HEC-RAS and SWMM are used to assess generated control schedules \cite{sadler2019leveraging, leon2014dynamic, yin2023physic}. 
However, such methods are prohibitively slow since they require thousands of time-consuming physics-based simulations \cite{jafarzadegan2023recent}.

Over the past decade, machine learning (ML) has become an increasingly powerful tool in various aspects of flood management, including flood prediction \cite{shi2023deep}, flood detection \cite{tanim2022flood}, susceptibility assessment \cite{saha2021flood}, and post-flood management \cite{munawar2019after}. 
Despite these advances, ML-based methods have yet to be widely applied to flood mitigation tasks.
We address this gap by introducing a fully machine learning-based framework, comprising a \emph{generator} to generate control schedules and a \emph{simulator} to assess them. 
The ML-based framework offers significantly faster response times than compute-intensive physics-based approaches, enabling real-time flood management.
Furthermore, the proposed ML-based method allows the simulator to actively guide the generator through gradient descent, thus seamlessly integrating the generator and simulator components and leveraging the differentiability of the learned simulator model \cite{sv2023gradient}.
To this end, we propose \underline{\textbf{F}}orecast \underline{\textbf{I}}nformed \underline{\textbf{D}}eep \underline{\textbf{L}}earning \underline{\textbf{Ar}}chitecture, \FIDLAr, for flood management. \FIDLAr's characteristics are summarized as follows:
\begin{itemize}
    \item \FIDLAr\ seamlessly combines two DL models in series: \texttt{Flood Manager} and \texttt{Flood Evaluator}.
    The former model is responsible for generating water pre-release schedules, while the latter model accurately forecasts the resulting water levels. Moreover, with the gradient-based planning, and differentiability of trained \texttt{Evaluator}, it can reinforce the \texttt{Manager} to generate better schedules.
    \item \FIDLAr\ is a data-driven approach, learning flood mitigation strategies from historically observed data.
    Once trained, it offers rapid response capabilities, highlighting the advantages of DL-based models over physics-based models, particularly for real-time flood management.
    \item \FIDLAr\ is a model-agnostic framework, where both the \texttt{Manager} and \texttt{Evaluator} could be any type of DL model trained with differentiable loss functions that allow back-propagation.
    \item \FIDLAr\ is trained with a customized loss function to balance flood risks and water wastage at the same time, which is a multitask learning method.
\end{itemize}

\section{Related Work}
\label{sec:relate}
\paragraph{Flood Prediction.} Physics-based mechanic models (e.g., HEC-RAS, SWMM) have been widely used to simulate water levels and flows in river systems \cite{peker2024integration,rahman2024drivers,gomes2023modeling,rivett2022acute}. 
However, these models are computationally inefficient and fall short of capturing precise knowledge of study domains \cite{bentivoglio2022deep}. Therefore, diverse machine learning (ML) and deep learning (DL) models have been studied as surrogates to simulate water levels and flows.
For example, support vector machines (SVMs) have been used to predict the urban flash floods \cite{yan2018urban,choubin2019ensemble}, multivariate regression models were adopted to estimate flood volumes and peak flows \cite{yang2020regional}, random forests and K-nearest neighbors were explored for urban flood inundation mapping \cite{castro2014flood}, gaussian process learning models for fast and accurate flood inundation simulation \cite{fraehr2023development}. Furthermore, deep learning models, such as recurrent neural networks, convolutional neural networks, and Transformers, have been employed for flood inundation \cite{zhou2021rapid} and flood prediction \cite{shi2023graph,shi2023deep}.

\paragraph{Flood Mitigation.} Flood mitigation seeks control schedules of those hydraulic structures to avoid or mitigate flood risks, which requires flood prediction models as simulators to evaluate the control schedules.
Researchers have attempted to leverage the genetic algorithm and the pattern search to generate control schedules and physics-based models (e.g., EPA-SWMM5, HEC-RAS) as water simulators \cite{sadler2019leveraging, leon2014dynamic, leon2020matlab}. The Lake Mendocino Operations (LMO) model was developed to simulate operations of Lake Mendocino such as release constraints for flood control and water supply operations \cite{delaney2020forecast}.
However, such methods are computationally intensive since they require thousands of time-consuming simulation trials with physics-based models \cite{jafarzadegan2023recent}.
Additionally, the genetic algorithm and pattern search techniques are usually heuristic, without positive feedback or guidance to better control those hydraulic structures.
\section{Problem Formulation}
\label{sec:problem}
Flood management or mitigation aims to manage water levels before extreme weather events. 
It involves predicting control schedules for hydraulic structures such as gates and pumps within the river system, denoted as $X^{gate, pump}_{t+1:t+k}$, spanning $k$ time points into the future from $t+1$ to $t+k$.
This prediction takes as input historical data on all possible variables (see Figure \ref{fig:domain}), $X$, from the preceding $w$ time points, in conjunction with reliably forecasted covariates (such as rainfall and tide) for the next $k$ time points.
Then we could train a deep learning (DL) model, $\mathcal{M}_{\theta}$, with parameters $\theta$:
\begin{equation}
\label{eq:problem_formulation}
    \mathcal{M}_{\theta_M}: (X^{all}_{t-w+1:t}, X^{cov}_{t+1:t+k}) \rightarrow  X^{gate, pump}_{t+1:t+k},
\end{equation}
where the subscripts represent the time ranges, and the superscripts refer to the variables under consideration.
Superscripts are dropped when all variables are taken into account. The superscript $cov$ refers specifically to the reliably predicted covariates (e.g., rain, tides). 
\section{Methodology}
\label{sec:method}

\subsection{Overview}
\label{sec:challenge}
Intuitively, an ML model can be trained to learn the function $\mathcal{M}_{\theta_M}$.
However, a key challenge lies in the historical data, which often reflects control schedules that led to flooding or other suboptimal outcomes, making it unsuitable as ground-truth data for traditional supervised learning.
To overcome this, we train the model, $\mathcal{M}_{\theta_M}$, by leveraging the differentiability of a learned simulator model, $\mathcal{E}_{\theta_E}$.
First, we pre-train an independent and accurate \texttt{Flood Evaluator} using extensive historical data to model the \emph{consequences} (e.g., water levels) of various \emph{actions} (e.g., control schedules).
We then frame the control schedule planning in \texttt{Flood Manager} as an optimization problem, seeking actions that minimize undesirable outcomes - floods or water wastage. Both the \texttt{Evaluator} and \texttt{Manager} are implemented as neural networks, with the framework illustrated in Figure \ref{fig:fidla_framework}.

\subsection{Flood Evaluator}
\label{sec:evaluator}
\texttt{Flood Evaluator} is tasked with accurately forecasting water levels at designated points of interest within river systems for any control schedule of gates and pumps.
The underlying transfer function of the \texttt{Evaluator} is:
\begin{equation}
\label{eq:WaLeF}
    \mathcal{E}_{\theta_E}: (X^{all}_{t-w+1:t}, X^{cov}_{t+1:t+k}, X^{gate, pump}_{t+1:t+k}) \rightarrow X^{water}_{t+1:t+k}.
\end{equation}
\noindent The \texttt{Evaluator} is trained independently using large-scale historical data to achieve highly accurate water level predictions for any given set of conditions and control schedules.
Therefore, once the \texttt{Evaluator} is well trained, its parameters are frozen while training the \texttt{Manager}, where it plays the role of a trained ``referee'' - scoring control schedules generated by the \texttt{Manager} by predicting the resulting water levels. 
It also serves to backpropagate the gradient descent feedback, guiding the \texttt{Manager} to produce more effective control schedules of gates and pumps.

\begin{figure}[ht]
\centering
    \includegraphics[width=0.95\columnwidth]{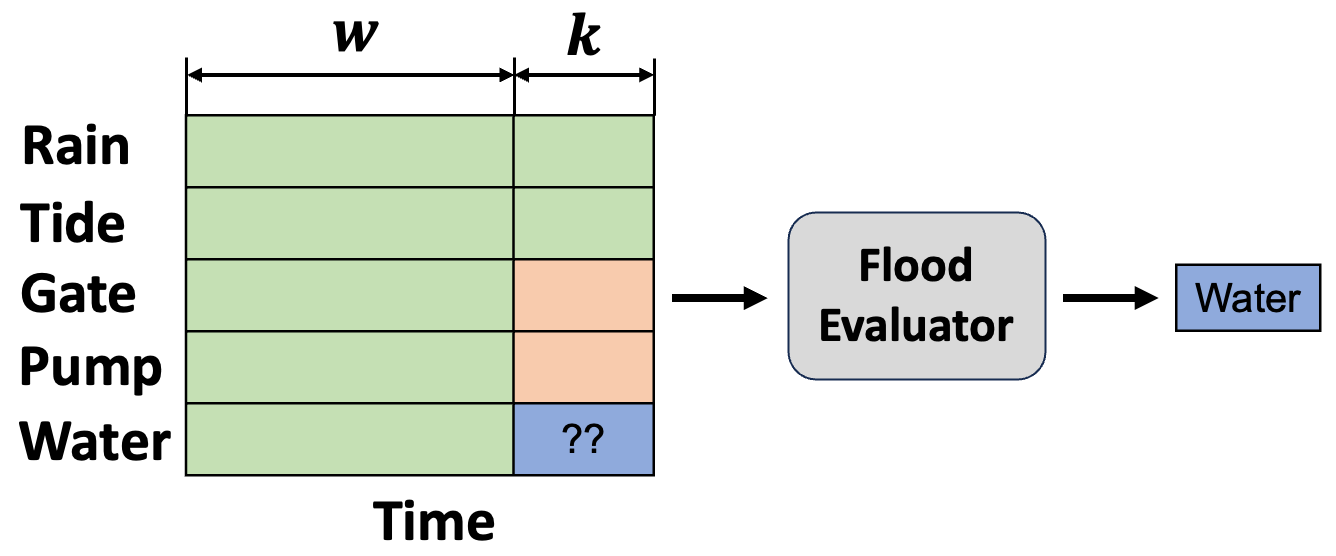} 
\caption{{\bf Flood Evaluator.} The parts shaded green are used as inputs. 
Water levels (blue) are the outputs.}
\label{fig:flood_evaluator}
\end{figure}

\begin{algorithm}[ht!]
\small
\caption{Training algorithm of \FIDLAr}
\label{alg:train_algorithm}
\textbf{Input}: recent past data: ${\bf X}^{all}_{t-w+1, t}$\\ 
\textcolor{white}{\textbf{Input}:} near future data: ${\bf X}^{cov}_{t+1, t+k} = {\bf X}^{rain, tide, gate, pump}_{t+1, t+k}$ \\
\textbf{Parameter}: $\theta_E, \theta_M$: parameters of Evaluator and Manager; $w, k:$ length of past and prediction windows
\begin{algorithmic}[1] 
    \STATE \textcolor{blue}{// \texttt{Train Flood Evaluator,} $\mathcal{E}_{\theta_E}$}
    \STATE initialize learnable parameters $\theta_E$
    \FOR{$i = 1, \ldots, N$ epochs}
        \STATE $\text{MiniBatch} \gets (\{{\bf X}^{all}_{t-w+1, t}, {\bf X}^{cov}_{t+1, t+k}\}, {\bf X}^{water}_{t+1, t+k})$
        \STATE $\hat{{\bf X}}^{water}_{t+1, t+k} \gets \mathcal{E}_{\theta_E}({\bf X}^{all}_{t-w+1, t}, {\bf X}^{cov}_{t+1, t+k})$
        \STATE $\mathcal{L}_{E} \gets \frac{1}{k}\sum_{j=1}^k||\hat{{\bf X}}^{water}_{j} - {\bf X}^{water}_{j}||^2$
        \STATE $\nabla_{\theta_E} \gets \text{BackwardAD}(\mathcal{L}_{E})$
        \STATE $\theta_{E} \gets \theta_{E} - \eta \nabla_{\theta_E}$
    \ENDFOR
    \STATE \textbf{return} trained \texttt{Flood Evaluator}, $\mathcal{E}_{\theta_E}$
    \STATE \textcolor{blue}{// \texttt{Train Flood Manager,} $\mathcal{M}_{\theta_M}$, with frozen $\mathcal{E}_{\theta_E}$}
    \STATE initialize learnable parameters $\theta_M$
    \WHILE{$X^{water}_{t, t+k}$ violates either threshold}
        \STATE $\text{MiniBatch} \gets (\{{\bf X}^{all}_{t-w+1, t}, {\bf X}^{rain, tide}_{t+1, t+k}\}, {\bf X}^{gate, pump}_{t+1, t+k})$
        \STATE $\hat{{\bf X}}^{gate, pump}_{t+1, t+k} \gets \mathcal{M}_{\theta_M}({\bf X}^{all}_{t-w+1, t}, {\bf X}^{rain, tide}_{t+1, t+k})$
        \STATE $\hat{{\bf X}}^{water}_{t+1, t+k} \gets \mathcal{E}_{\theta_E}({\bf X}^{all}_{t-w+1, t}, {\bf X}^{rain, tide}_{t+1, t+k}, \hat{{\bf X}}^{gate, pump}_{t+1, t+k})$ 
        \STATE $\mathcal{L}_{E} = c_1 \cdot \mathcal{L}_{1}(\hat{{\bf X}}^{water}_{t+1, t+k}) + c_2 \cdot \mathcal{L}_{2}(\hat{{\bf X}}^{water}_{t+1, t+k})$ 
        \STATE $\nabla_{\theta_M} \gets \text{BackwardAD}(\mathcal{L}_{E})$
        \STATE $\theta_{M} \gets \theta_{M} - \eta \nabla_{\theta_M}$
    \ENDWHILE
    
    \STATE \textbf{return} trained \texttt{Flood Manager}, $\mathcal{M}_{\theta_M}$
\end{algorithmic}
\end{algorithm}

\subsection{Flood Manager}
\label{sec:manager}
\texttt{Flood Manager} is to produce control schedules for hydraulic structures (i.e., gates and pumps), taking as inputs reliably predictable future information (rain, tide) and all historical data.
Since no ground truth is available, it is trained with the differentiability of the learned \texttt{Evaluator} model.
Therefore, we connect the \texttt{Manager} with the \texttt{Evaluator} where the output of Eq. (\ref{eq:problem_formulation}) is injected into Eq. (\ref{eq:WaLeF}):
\begin{equation}
\label{eq:predicted_gate_in_WaLeF}
    \mathcal{E}_{\theta_E}(X^{all}, X^{cov}_{t+1:t+k}, \mathcal{M}_{\theta_M}(X^{all}, X^{cov}_{t+1:t+k})) \rightarrow X^{water}_{t+1:t+k},
\end{equation}
where $X^{all}=X^{all}_{t-w+1:t}$ and $\theta_M$ and $\theta_E$ are the parameters of \texttt{Manager} and \texttt{Evaluator}.

The resulting output of water levels can be used to compute the loss in Eq. (\ref{eq:total_loss}), representing evaluation scores for generated control schedules.
Gradient descent \cite{ruder2016overview} can be back-propagated as the feedback to update the parameters of the \texttt{Manager}. The parameter update is
presented:
\begin{equation}
    \theta_M := \theta_M - \alpha \cdot \frac{\partial \mathcal{L}}{\partial \theta_M},
    \label{eq:gradient}
\end{equation}
\noindent where $\alpha$ is the learning rate and $\frac{\partial \mathcal{L}}{\partial \theta_M}$ is the partial derivative of the compound function $\frac{\partial \mathcal{L}}{\partial \theta_M} = \frac{\partial \mathcal{L}}{\partial \mathcal{E}} \cdot \frac{\partial \mathcal{E}}{\partial \mathcal{M}} \cdot \frac{\partial \mathcal{M}}{\partial \theta_M}$.
The training details of \FIDLAr\ are given in Algorithm \ref{alg:train_algorithm}.
\begin{figure}[ht]
\centering
    \includegraphics[width=0.9\columnwidth]{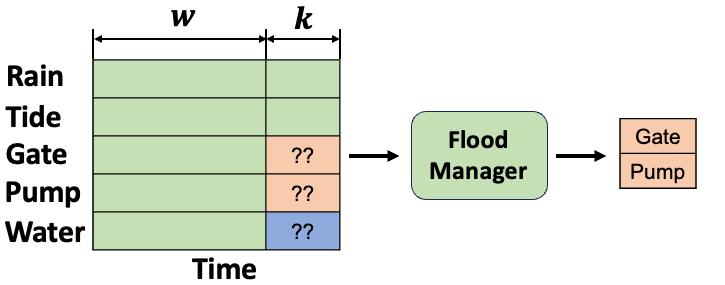}
\caption{{\bf Flood Manager.} 
The parts shaded green (historical data) are the inputs, and the parts shaded orange are the outputs. The water levels shaded blue are not predicted.}
\label{fig:flood_manager}
\end{figure}

\subsection{Custom Loss Function}
\label{sec:loss}
\begin{figure}[ht]
\centering
    \begin{subfigure}[b]{0.22\textwidth}
        \centering
        \includegraphics[scale=0.28]{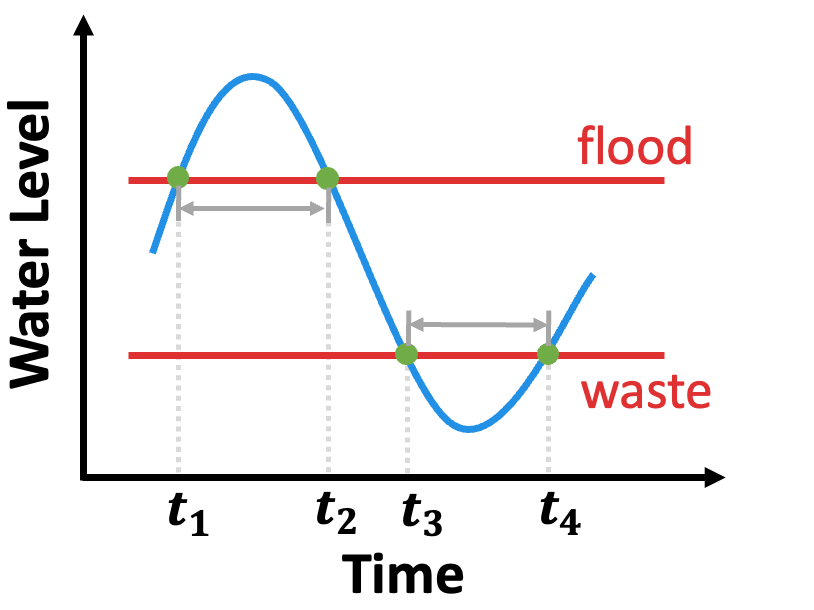}
        \caption{}
        \label{fig:loss_a}
    \end{subfigure}
    \hfill
    \begin{subfigure}[b]{0.23\textwidth}
        \centering
        \includegraphics[scale=0.28]{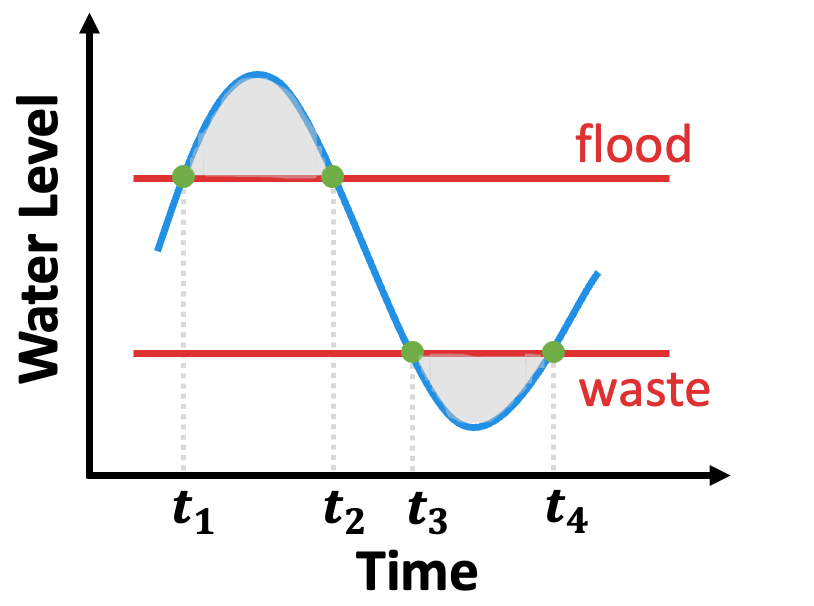}
        \caption{}
        \label{fig:loss_b}
    \end{subfigure}
\caption{The two red bars represent a threshold of flooding and a threshold of water wastage.
Shown are (a) the time spans when these thresholds are crossed, and (b) the areas between water level curves and threshold bars.
Violations of the upper and lower thresholds are captured in $\mathcal{L}_1$ and $\mathcal{L}_2$.}
\label{fig:loss}
\end{figure}

Loss functions are critical in steering the learning process. 
An obvious one is the total time (Figure \ref{fig:loss_a}) for which the water levels either exceed the \textit{flooding threshold} or dip below the \textit{water wastage threshold}.
Another related metric is the extent to which the limits are exceeded to signify the severity of floods or water wastage (Figure \ref{fig:loss_b}).
The lower threshold for flood management is important in practice, since it prevents water wastage, thereby supporting irrigation, facilitating navigation, and maintaining ecological balance.
It also prevents the optimization methods from trivially recommending the depletion of valuable water resources to prevent future flooding. 
The $\mathcal{L}_1$ and $\mathcal{L}_2$ represent the \textit{flooding} and \textit{water wastage} losses, respectively, and the final loss function is a balanced combination as shown in Eq. (\ref{eq:total_loss}).
\begin{equation}
\label{eq:flood_loss}
    \begin{aligned}
        \mathcal{L}_{1} = \sum^{N}_{i=1} \sum^{t+k}_{j=t+1} \Arrowvert max\{\hat{X}^{water}_{i, j} - X^{flood}_{i}, 0\} \Arrowvert^{2}, \\ 
        \mathcal{L}_{2} = \sum^{N}_{i=1} \sum^{t+k}_{j=t+1} \Arrowvert min\{\hat{X}^{water}_{i, j} - X^{waste}_{i}, 0\} \Arrowvert^{2},
    \end{aligned}
\end{equation}
where $N$ is the number of water level locations of interest; $k$ is the length of prediction horizon; $X^{flood}$ and $X^{waste}$ represent the thresholds for flooding and water wastage; and the capped version, $\hat{X}^{water}$, is obtained using the \texttt{Evaluator} module.
The combined loss function is given by:
\begin{equation}
\label{eq:total_loss}
    \mathcal{L}_{total} = c_1 \cdot \mathcal{L}_1 + c_2 \cdot \mathcal{L}_2,
\end{equation}
where $c_1/c_2$ dictates the relative importance of $\mathcal{L}_1$ and $\mathcal{L}_2$.

\section{Graph Transformer Network}
The \texttt{Manager} and \texttt{Evaluator} modules described so far are model agnostic. We tried many existing architectures for them, as discussed in Section \ref{sec:exp}.
We devise the \underline{G}raph \underline{T}ransformer \underline{N}etwork (GTN) architecture by combining graph neural networks (GNNs), attention-based transformer networks, long short-term memory networks (LSTMs), and convolutional neural networks (CNNs).
GNN and LSTM modules are combined to learn the spatiotemporal dynamics of water levels, while the Transformer and CNN modules focus on extracting feature representations from the covariates. 
The \emph{attention} mechanism \cite{vaswani2017attention} is used to discern interactions between covariates and water levels, as shown in Eq. (\ref{eq:attention}) below.
Figure \ref{fig:graphtransformer} presents the GTN architecture, which is used for both \texttt{Evaluator} and \texttt{Manager}, but with minor changes accordingly of the inputs and outputs (see Figures \ref{fig:flood_evaluator} and \ref{fig:flood_manager}). We have two GNN layers with 32 and 16 channels, one LSTM layer with 16 units, one CNN with 96 filters, and one Transformer encoder with 3 heads.
\begin{equation}
\label{eq:attention}
    \begin{aligned}
    Atte(Q, K, V) {} & = softmax(\frac{Q^{cov} (K^{water})^T}{\sqrt{d}}) V^{water}  \\
                     & = softmax(\frac{Q^{water} (K^{cov})^T}{\sqrt{d}}) V^{cov},  \\
    \end{aligned}
\end{equation}
where $T$ denotes the transpose operation; $water$ and $cov$ represent water levels and covariates; and $d$ is the length of projection embedding where $d=d_q=d_k=d_v=128$.

\section{Experiments}
\subsection{Dataset}
\begin{figure}[ht!]
\centering
    \includegraphics[width=0.93\columnwidth]{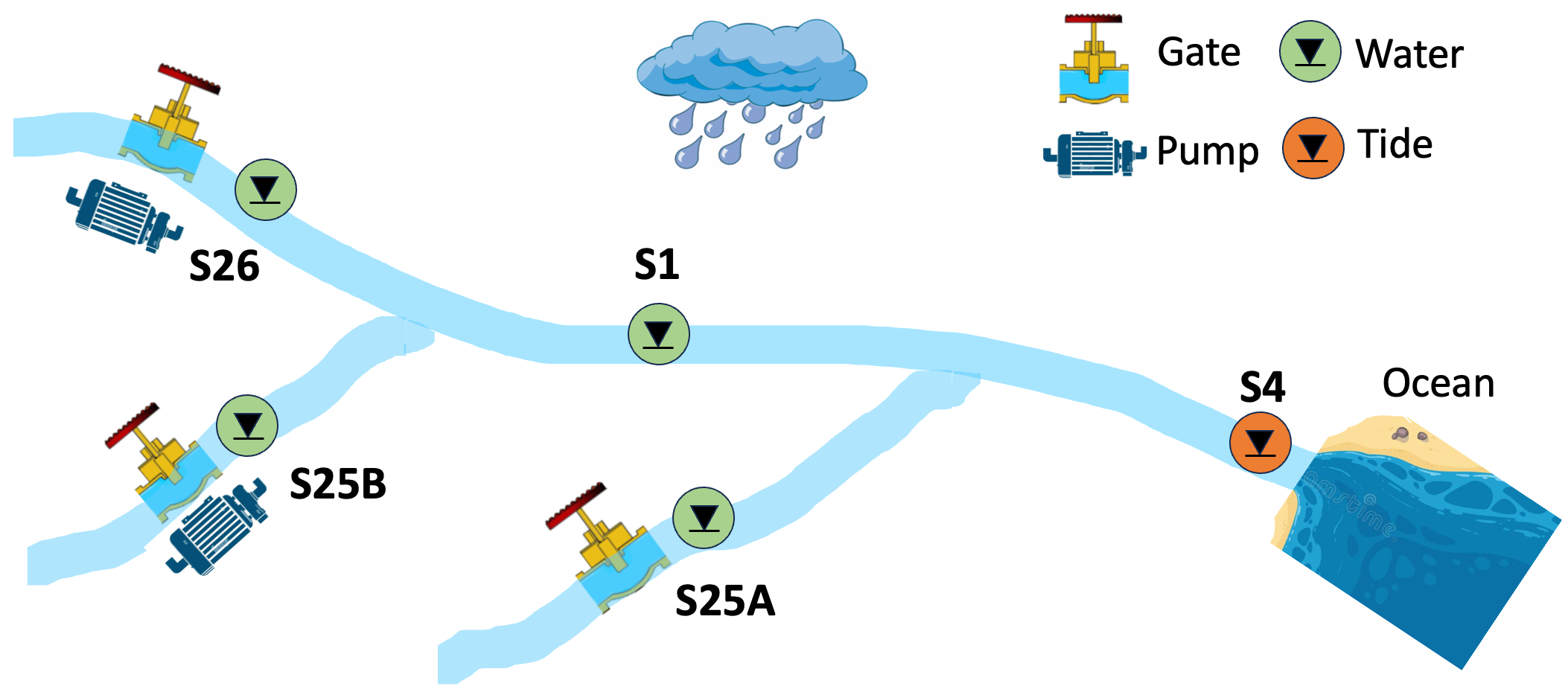} 
\caption{Schemetic diagram of study domain. There are three water stations with hydraulic structures (S26, S25B, S25A), one simple water station, S1 (green circle in the middle), and a station monitoring the tide level from the ocean.}
\label{fig:domain}
\end{figure}

%
%
We obtained data from the South Florida Water Management District's (SFWMD) DBHydro database \cite{dbhydro2023sfwmd} for the coastal stretch in South Florida.
The data set consists of hourly observations for water levels and external covariates from January 1, 2010 to December 31, 2020.
As shown in Figure \ref{fig:domain}, the river system has two branches and includes several hydraulic structures (gates, pumps) to control water flows.
We aim to predict effective control schedules on hydraulic structures (gates, pumps) to minimize flood risks at four specific locations marked by green circles.
\begin{table}[ht!]
\centering
\setlength{\tabcolsep}{4pt} 
    \begin{tabular}{lcccc}
        \toprule
        \textbf{Feature}    & \textbf{Interval}  & \textbf{Unit}  & \textbf{\#Var.} & \textbf{Location}  \\
        \midrule \midrule    
        Rainfall            &Hourly         & $inch/h$      & 1    & -  \\ 
        Tide                &Hourly         & $ft$          & 1    & S4  \\ 
        Pump                &Hourly         & $ft^3/s$      & 2    & S25B, S26  \\ 
        Gate                &Hourly         & $ft$          & 3    & S25A, S25B, S26  \\
        Water               &Hourly         & $ft$          & 4    & S25A, S25B, S26, S1  \\    
        \bottomrule 
    \end{tabular}
\caption{Summary of the data set.}
\label{tab:data_summary}
\end{table}
\begin{figure*}[ht]
\centering
\includegraphics[width=1.8\columnwidth]{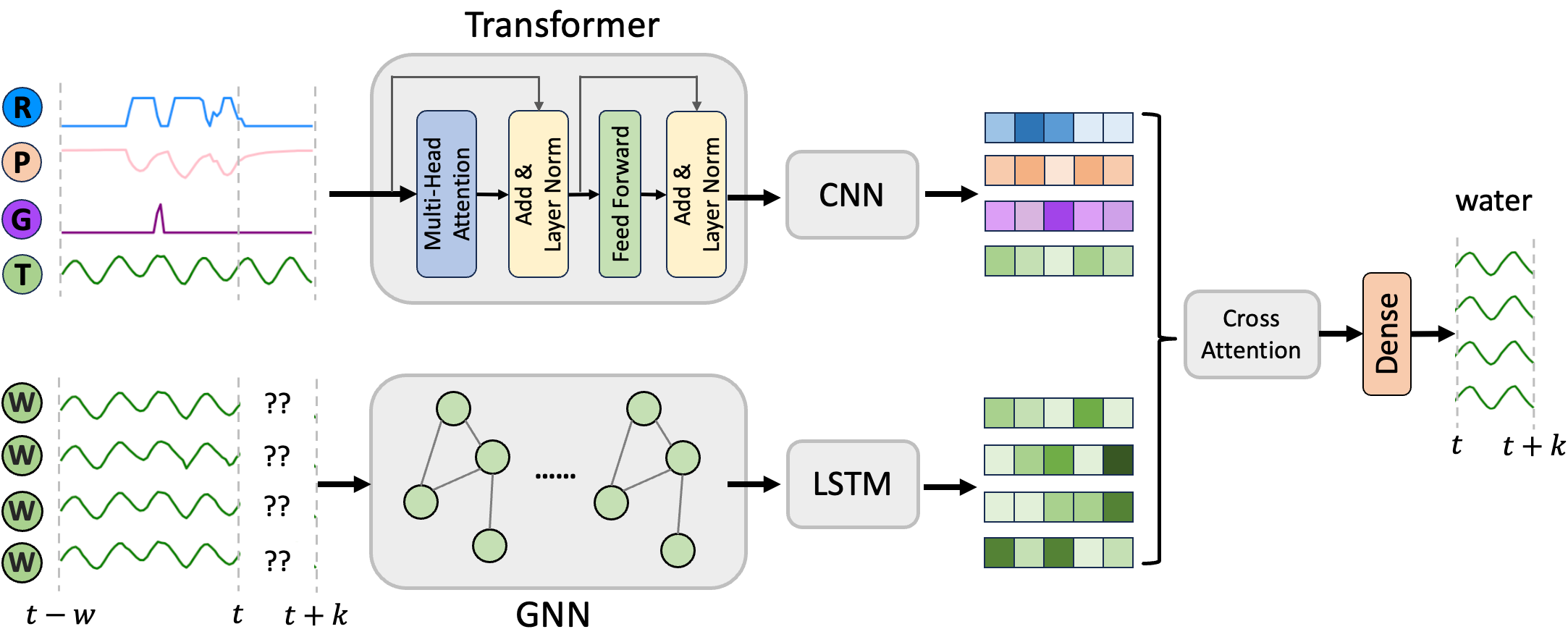} 
\caption{Graph Transformer Network for \texttt{Flood Evaluator}. Input variables include Rainfall, Pump, Gate, Tide, and Water levels as shown in Figure \ref{fig:domain} and Table \ref{tab:data_summary}, which are generically denoted by $R, T, G, P$, and $W$. Output is water levels.}
\label{fig:graphtransformer}
\end{figure*}
%
%
\subsection{Experimental Design}
\label{sec:exp}
The sliding input window \cite{li2014rolling} (also known as look-back window \cite{gidea2018topological} strategy was used to process the entire dataset \cite{shi2023explainable}). For consistency, we used a look-back window of length $w=72$ hours and a prediction window of length $k=24$ hours.
The dataset was split in chronological order with the first 80\% for training and the remaining 20\% for testing. 
The eight DL methods below are used for \texttt{Flood Manager} and \texttt{Flood Evaluator}. 
We run all experiments on one NVIDIA A100 GPU with 80GB memory.
\begin{itemize}
\item \textbf{MLP} \cite{suykens1995artificial}: Multilayer perceptron can learn non-linear dependencies; 
\item \textbf{RNN} \cite{medsker2001recurrent}: Recurrent neural networks are good at processing sequential data;
\item \textbf{CNN} \cite{o2015introduction}: A 1D convolutional neural network;
\item \textbf{GNN} \cite{kipf2016semi}: Graph neural network with nodes representing variables and edges representing spatial dependencies;
\item \textbf{TCN} \cite{bai2018empirical}: Temporal dilated convolutional network with an exponentially large receptive field;
\item \textbf{RCNN} \cite{zhang2020temperature}: Combined RNN and CNN model for time series forecasting;
\item \textbf{Transformer} \cite{vaswani2017attention}: Attention-based network for sequence modeling (only encoder);
\item \textbf{GTN} (Ours): Combining GNNs with LSTMs, CNNs, and transformers, as described in Figure \ref{fig:graphtransformer}. 
\end{itemize}

\subsubsection{Flood Prediction.} 
The role of the \texttt{Flood Evaluator} is to forecast flood events by predicting water levels for given input conditions. 
We set the upper threshold (flood level) at 3.5 feet and the lower threshold (wastage level) at 0.0 feet.
However, the methods remain consistent for many reasonable choices of threshold values. 
We measured accuracy using multiple metrics: (a) mean absolute error (MAE), (b) root mean squared error (RMSE) computed between the predicted and actual water levels, (c) number of time points where the upper or lower thresholds are breached, and (d) the area between water level curves and threshold bars.
Table \ref{tab:flood_prediction_s1} demonstrates that our model \texttt{GTN} outperforms other models with predictions (in \textcolor{red}{red}) most closely aligned with the ground truth (in \textcolor{blue}{blue}) while achieving the lowest MAE and RMSE (in \textcolor{orange}{orange}). 
Therefore, we choose our GTN model as \texttt{Evaluator} while training \texttt{Manager} in \FIDLAr.
In the Appendix, we provide more results in Table \ref{tab:flood_prediction}.
\begin{table*}[ht]
\small
\centering
    \begin{tabular}{l|cc|cccc}
    \toprule
    \textbf{Methods}    & \textbf{MAE (ft)}   & \textbf{RMSE (ft)} & \textbf{Over Timesteps}   & \textbf{Over Area}  & \textbf{Under Timesteps}   & \textbf{Under Area} \\
    \midrule\midrule
    Ground-truth      & -            & -        & \textcolor{blue}{96}      & \textcolor{blue}{14.82}     & \textcolor{blue}{1,346}    & \textcolor{blue}{385.80}  \\ 
    \midrule
    HEC-RAS           & 0.174        & 0.222    & 68       & 10.07    & 1,133    & 325.33  \\ 
    \midrule
    MLP               & 0.065        & 0.086    & 147      & 27.96    & 1,677    & 500.41  \\  
    RNN               & 0.054        & 0.072    & 110      & 17.12    & 1,527    & 441.41  \\   
    CNN               & 0.079        & 0.104    & 58       & 5.91     & 1,491    & 413.22  \\ 
    GNN               & 0.054        & 0.070    & 102      & 15.90    & 1,569    & 462.63  \\ 
    TCN               & 0.050        & 0.065    & 47       & 5.14     & 1,607    & 453.63  \\  
    RCNN              & 0.092        & 0.110    & 37       & 4.61     & 1,829    & 553.20  \\
    Transformer       & 0.050        & 0.066    & 151      & 25.95    & 1,513    & 434.13  \\ 
    \midrule
    GTN (ours)   & \textcolor{orange}{0.040}        & \textcolor{orange}{0.056}    & \textcolor{red}{100}      & \textcolor{red}{15.64}    & \textcolor{red}{1,390}    & \textcolor{red}{398.84}  \\  
    \bottomrule
    \end{tabular}
\caption{Comparison of model performances for the \texttt{Flood Evaluator} on the test set, specifically at time t+1 for measurement station S1. The terms ``Over Timesteps'' and ``Under Timesteps'' indicate the number of time steps during which water levels exceed the upper threshold or fall below the lower threshold, respectively. Similarly, ``Over Area'' and ``Under Area'' pertain to the area between the water level curve and upper or lower threshold, as was illustrated in Figure \ref{fig:loss}. Results in \textcolor{orange}{orange} are the lowest in that column while results in \textcolor{red}{red} are the closest to the ground truth (in \textcolor{blue}{blue}).}
\label{tab:flood_prediction_s1}
\end{table*}

\subsubsection{Flood Mitigation with \FIDLAr.} 
\FIDLAr\ requires both \texttt{Evaluator} and \texttt{Manager} components.
For the \texttt{Manager} model, we experimented with one rule-based method, and two genetic algorithms -- one with a physics-based HEC-RAS evaluator \cite{leon2020matlab} and one with our DL-based GTN evaluator, and several DL-based managers using MLP, RNN, CNN, GNN, TCN, RCNN, Transformer, and GTN.
\FIDLAr\ was measured using (a) the number of time steps where the upper/lower thresholds are exceeded for the water levels, and (b) the area between the water level curves and the threshold bars.
Table \ref{tab:flood_mitigate_s1} shows that all DL-based methods consistently performed better for site S1 than rule-based and GA-based approaches. 
Furthermore, GTN has the best performance under all four metrics, whether it is to control floods or water wastage.
The results for all other measurement sites follow a similar pattern, as shown in Table \ref{tab:flood_mitigate} in Appendix \ref{sec:more_exp_flood_miti}.
\begin{table*}[ht]
\small
\centering
    \begin{tabular}{l|c|cccc}
    \toprule
    \multirow{1}{*}{\textbf{Method}}
        & \textbf{Manager}        & \textbf{Over Timesteps}    & \textbf{Over Area}   & \textbf{Under Timesteps}    & \textbf{Under Area}  \\
    \midrule \midrule 
    \multirow{1}{*}{Rule-based}
        &                   & 96         & 14.82     & 1,346     & 385.8  \\  
    \midrule
    \multirow{2}{*}{GA-based}    
        & Genetic Algorithm$^{*}$     & -          & -         & -         & -  \\
        & Genetic Algorithm$^{\dag}$     & 86         & 16.54     & 454       & 104  \\
    \midrule
    \multirow{8}{*}{DL-based}       
        & MLP                       & 91         & 13.31     & 1,071     & 268.35 \\  
        & RNN                       & 35         & 3.97      & 351       & 61.05  \\   
        & CNN                       & 81         & 11.22     & 1,163     & 314.37  \\ 
        & GNN                       & 31         & 3.72      & 429       & 84.31  \\ 
        & TCN                       & 39         & 3.77      & 306       & 55.12  \\  
        & RCNN                      & 29         & 3.28      & 328       & 58.68  \\
        & Transformer               & 85         & 11.54     & 1,180     & 310.16  \\ 
        & GTN (Ours)                &\textbf{22} &\textbf{2.23} &\textbf{299}  &\textbf{53.34}  \\
    \bottomrule
    \end{tabular}
\caption{Comparison of model performances for the \texttt{Flood Manager} on the test set, specifically at time t+1 for measurement station S1. The $^{*}$ denotes that the GA method was used with a physics-based (HEC-RAS) evaluator. The $-$ denotes that the experiments were timed out. The $^{\dag}$ denotes the GA method was used with the GTN as the evaluator. All other rows are DL-based flood managers with a DL-based GTN as the evaluator. 
Results in bold are the best in that column.} 
\label{tab:flood_mitigate_s1}
\end{table*}

\begin{figure}[ht]
\centering
    \includegraphics[width=0.93\columnwidth]{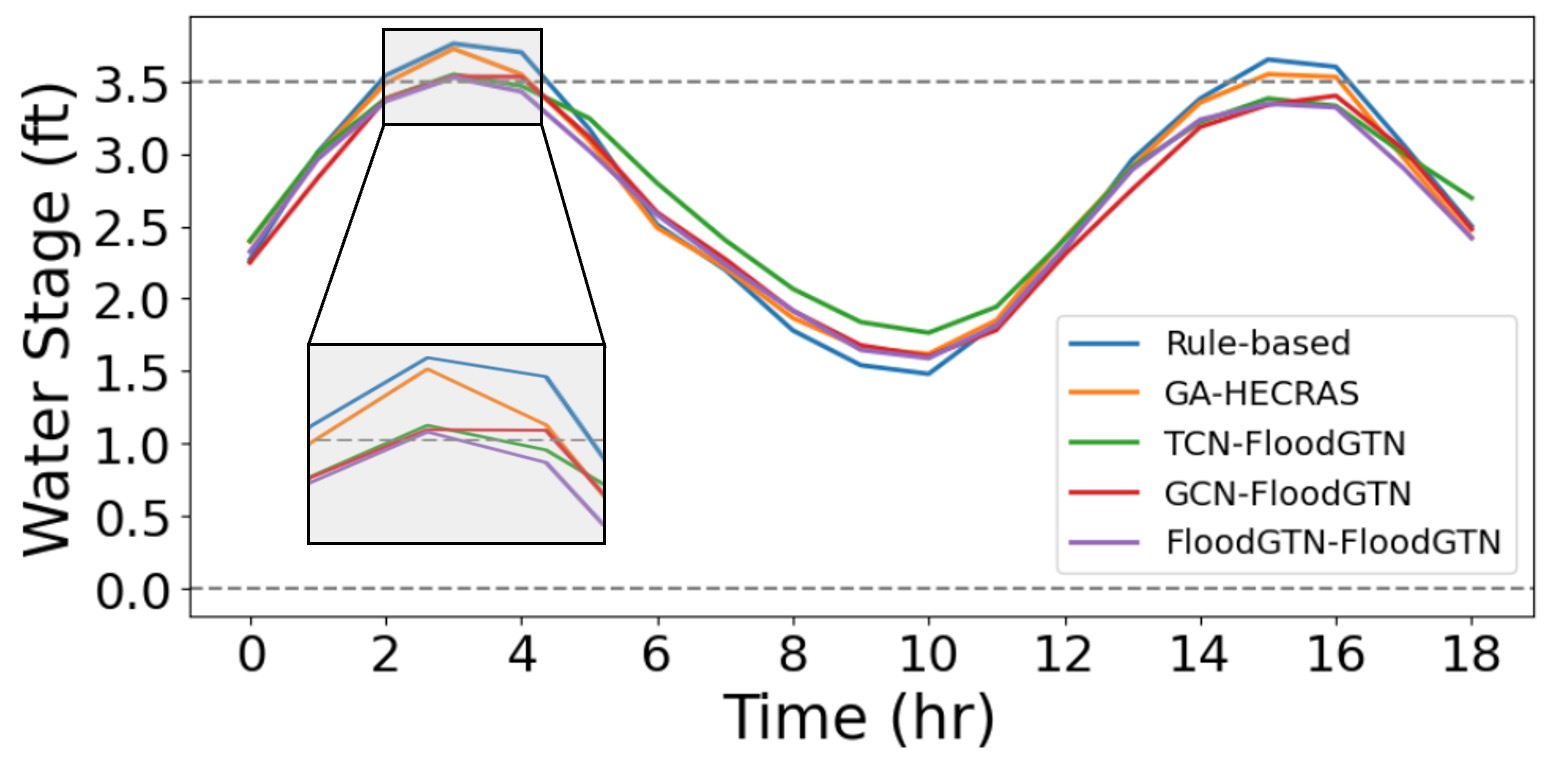} 
\caption{Visualization of water levels with various methods for flood mitigation. We zoomed in $t=2\sim4.5$ in gray. `A'--`B' in legend represents the \texttt{Manager} and \texttt{Evaluator}.}
\label{fig:visualize_flood_mitigation_s1}
\end{figure}

We visualize water levels for a short period spanning 18 hours from September 3rd (09:00) to September 4th (03:00) in 2019 for the S1 location.
Figure \ref{fig:visualize_flood_mitigation_s1} indicates that \FIDLAr\ equipped with GTN model (purple curve) has led to water levels within the upper and lower thresholds, satisfying pre-defined requirements.
Moreover, \FIDLAr\ presents the best control (i.e., the least water levels beyond thresholds) for flood mitigation and water waste compared to other baselines.
We zoomed in on a 2.5-hour period of the resulting decreased water levels. 
Its quantitative performance is in Table \ref{tab:flood_mitigate_event_s1} while more visuals are in Appendix \ref{sec:vis_appendix}.

\subsection{Ablation Study}
The \emph{ablation} study in Table \ref{tab:ablation_s1} quantifies the contribution of each component of GTN, by measuring the performance of GTN after removing each of the individual components.

\subsection{Analysis of Computational Time} 
Since \FIDLAr\ was designed for real-time flood control, we measured the running times of the models used in this work.
Table \ref{tab:flood_prediction_time} shows the running times for the whole flood prediction and mitigation system in its training and test phase. 
All the DL-based approaches in the test phase are several orders of magnitude faster than the currently used physics-based and GA-based approaches for the flood mitigation task. 
Rapid inference is a critical property of data-driven DL methods.
The table also shows the training times for the DL-based approaches, although they are not necessary for the deployment in reality.

\subsection{Explainability}
\label{sec:explain}
\begin{figure}[ht!]
\centering
    \includegraphics[width=0.98\columnwidth]{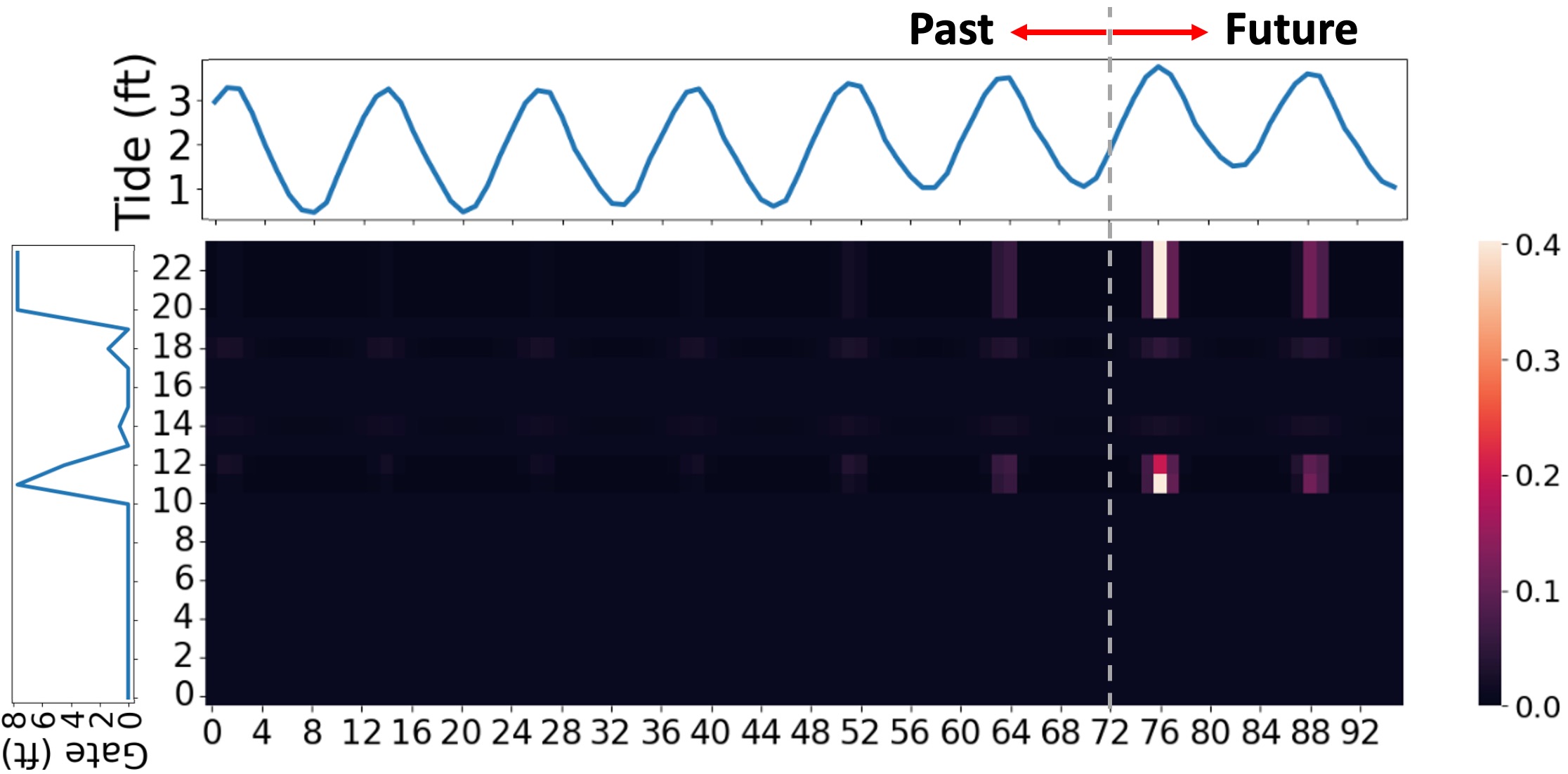} 
\caption{Importance scores of tide input. $x$ and $y$ axes are the tide and gate over time.}
\label{fig:gate_tide_attention}
\end{figure}

Attention-based methods allow us to calculate the ``attention scores'' assigned to an input variable to compute a specific output variable.
This is exemplified in the heatmap shown in Figure \ref{fig:gate_tide_attention}, which shows the attention scores assigned to the tide (columns) to compute the gate schedule output (rows) for 24 hours into the future.
Note that there are 96 columns and 24 rows because we use 72 hours of past tidal observations and 24 hours of future predicted tidal data to predict 24 hours of the gate schedule into the future.
The rows $[0,23]$ correspond to the 24 hours into the future, while the columns $[0,95]$ also include 72 hours of the recent past and 24 hours of the future predicted tidal data.
Therefore, $t=72$ corresponds to the ``current'' time point, and the columns $[72,95]$ correspond to the same time points as rows $[0,23]$.

\begin{table}[ht]
\small
\centering
    \begin{tabular}{l|cccc}
    \toprule
    \textbf{}   & \textbf{Over}    & \textbf{Over}   & \textbf{Under}    & \textbf{Under}  \\
    \textbf{Method}    & \textbf{Timesteps}    & \textbf{Area}   & \textbf{Timesteps}    & \textbf{Area}  \\
    \midrule \midrule     
     w/o CNN           & 37         & 4.37     & 476         & 85.54  \\  
     w/o Transformer   & 32         & 3.57     & 325         & 57.42  \\   
     w/o GNN           & 56         & 5.90     & 479         & 86.22  \\ 
     w/o LSTM          & 35         & 4.34     & 329         & 56.74  \\ 
     w/o Attention     & 32         & 3.59     & 341         & 60.48  \\  
    \midrule
    GTN (Ours)         &\textbf{22} & \textbf{2.23}  & \textbf{299}  & \textbf{53.34}  \\  
    \bottomrule
    \end{tabular}
\caption{Ablation study for flood mitigation for the entire test set (for time point t+1 at S1). The last row indicates the performance of the \FIDLAr\ system with GTN as proposed in Figure \ref{fig:graphtransformer}. The best results in the last row are in bold.}
\label{tab:ablation_s1}
\end{table}

\begin{table}[ht]
\small
\centering
\resizebox{\columnwidth}{!}{%
    \begin{tabular}{l|cccc}
        \toprule
        \multirow{2}{*}{\textbf{Model}}  & \multicolumn{2}{c}{\textbf{Flood Prediction}}        & \multicolumn{2}{c}{\textbf{Flood 
 Mitigation}} \\
                           & \textbf{Train}     & \textbf{Test}  & \textbf{Train}   & \textbf{Test} \\
        \midrule\midrule
        HEC-RAS            & -              & 45 min      & -              & -   \\ 
        Rule-based         & -              & -           & -              & -   \\
        GA$^{*}$           & -              & -           & -              & -     \\ 
        GA$^{\dag}$        & -              & -           & -              & est. 30 h  \\ 
        \midrule
        MLP                & 35 min         & 1.88 s      & 58 min         & 6.13 s \\
        RNN                & 243 min        & 8.57 s      & 54 min         & 12.75 s \\
        CNN                & 37 min         & 1.93 s      & 17 min         & 5.84 s \\
        GNN                & 64 min         & 3.13 s      & 29 min         & 7.26 s \\
        TCN                & 60 min         & 4.57 s      & 45 min         & 9.06 s \\
        RCNN               & 136 min        & 8.61 s      & 61 min         & 13.27 s \\
        Transformer        & 43 min         & 2.38 s      & 23 min         & 6.76 s \\
        GTN (Ours)         & 119 min        & 2.95 s      & 35 min         & 4.90 s \\
        \bottomrule
    \end{tabular}%
}
\caption{Running time for flood prediction and mitigation. The running time for the rule-based method is not reported since historical data was directly used. GA$^{*}$, which combines a GA-based tool and HEC-RAS, took too long and was not reported, although the time for a small data set is reported in the appendix of our arXiv version. 
GA$^{\dag}$, which combines the GA-based tool with GTN, also took too long but was estimated using a smaller sample.}
\label{tab:flood_prediction_time}
\end{table}

\section{Discussion}
\label{sec:concl}
\paragraph{Interpreting Attention Map.}
The explainability feature, which is shown with an example in Figure \ref{fig:gate_tide_attention}, can provide significant insights into our results.
Firstly, we point out that the brightest patches are in the last 24 columns of the heatmap.
Thus \FIDLAr\ pays greater attention to the 24 hours of future predicted tidal data than the past 72 hours, highlighting the importance of forecast information to a DL-based approach to flood mitigation.
While tides may have a more predictable pattern over time, the contribution of rain on the water levels can also be seen for other time points.
A second critical insight is that the brightest attention patches are in columns where the tide is at its highest is critical to the prediction of gate schedules.
Additionally, the water level at the first high tide peak after the ``current'' time is more significant than the other two.
Third, the gate schedule has peaks at times $t=11$ hour and $t=22$ hour into the future, which correspond to the lowest points of the tide.
This implies that the optimal time for pre-releasing water is during low tide phases. 
Opening gates during high tide periods in coastal river systems is less advisable, as it may lead to water flowing back upstream from the ocean.
Finally, we observe that there is a light patch around column $t=65$ hour, suggesting mild attention for the previous high tide peak, but almost no attention to any of the peaks before that.
This again suggests that we could have chosen to use a smaller window for the past input.
Doing this analysis could provide evidence for the right value of $w$, the size of the look-back window.

\section{Conclusions}
In this work, we present the shortcomings of the current approaches for flood mitigation and propose \FIDLAr, a DL-based tool to address the problem. \FIDLAr\ can compute water ``pre-release'' schedules for hydraulic structures in a river system to achieve effective and efficient flood mitigation, while ensuring that water wastage is avoided.
This was made possible by the use of well-crafted loss functions for the DL models.
The dual component design (with a \texttt{Manager} and an \texttt{Evaluator}) is a strength of \FIDLAr. 
It exploits the gradient-based planning and the differentiability of the trained \texttt{Evaluator} model for better optimization.
During training, the gradient-based back-propagation from the \texttt{Evaluator} helps to reinforce the \texttt{Manager}.

All the DL-based versions of \FIDLAr\ are several orders of magnitude faster than the (physics-based or GA-based) competitors while achieving improvement over other methods in flood mitigation. 
These characteristics allow us to entertain the possibility of real-time flood management, which was challenging for previous approaches.

\section*{Broader Impact}
This research focuses on flood mitigation which has an essential influence on disaster management.
As an AI application for social good, our model provides detailed responses on the management of hydraulic structures so as to avoid or mitigate flood events while ensuring minimal water wastage.
Our model can also serve as a warning tool. It helps local governments, emergency agencies, and residents to plan well in advance to minimize the losses and dangers caused by impending floods.

\section*{Acknowledgments}
This work is part of the Institute for Geospatial Understanding through an Integrative Discovery Environment (I-GUIDE) project, which is funded by the National Science Foundation under award number 2118329. Any opinions, findings, and conclusions or recommendations expressed herein are those of the authors and do not necessarily represent the views of funding sources.

\bibliography{references}

\newpage
\appendix
\onecolumn

\section{Data and Processing}
\label{sec:appendix_data}
\paragraph{Data.} 
We describe more details about the data set used in our work. 
This dataset includes water level measurements from multiple stations, control schedules for various hydraulic structures (such as gates and pumps) along the river, tide information, and rainfall data in South Florida, USA. 
We aim to predict effective schedules on hydraulic structures (gates, pumps at S25A, S25B, S26) to minimize flood risks at four specific locations as represented in the green circle in Figure \ref{fig:domain}.
The relative water stage $\in \text{[-1.25, 4.05]}$ feet. 
This dataset spans 11 years, ranging from 2010 to 2020, with hourly measurements. 
The primary feature description and diagram of the study domain are presented in Table \ref{tab:data_summary}.

\paragraph{Processing.}
\texttt{Flood Evaluator} is used to predict the water levels given the input of past $w$ time steps and some information of future $k$ time steps, either predicted information (``Rain'' and  ``Tide'') or pre-determined information (``Gate'' and ``Pump''). 
It serves as an evaluator to assess the gate and pump schedules by predicting the resulting water levels. \texttt{Flood Manager} is used to predict the gate and pump schedules given the input of past $w$ time steps and some information of future $k$ time steps, either predicted information (``Rain'' and  ``Tide''). 
Both models are trained as supervised learning tasks. Therefore, we pre-processed the data into pairs consisting of the input and output described above. 
In our experiments, we set $w=72$ hours and $k=24$ hours. The shape of the input and output is $(96, N_{input})$ and $(24, N_{output})$, where $N_{input}$ and $N_{output}$ are the number of input and output variables, respectively. 
Among the input, the future data with question marks is masked to a $mask\_value=1e-10$ using Keras API\footnote{\url{https://keras.io/api/layers/core_layers/masking/}}.
If all values in the input tensor at that time steps are equal to $mask\_value$, then the time steps will be masked (skipped) in all downstream layers.

\section{Model Architecture and Training Hyper-parameter}
\label{sec:model_details_app}
\subsection{Architecture}
\label{sec:appendix_arch}
Since FIDLAR is a model-agnostic framework, we use different deep learning models below as the backbone. In this section, we provide the model architecture, accordingly.
\begin{itemize}
\item \textbf{MLP}: We stacked two \texttt{Dense} layers with $64$ and $32$ neurons. 
Each \texttt{Dense} layer uses \texttt{RELU} as the activation function and $L_1$ and $L_2$ as the regularization factor. 
Each \texttt{Dense} layer is followed by a \texttt{Dropout} layer to avoid possible \textit{overfitting}.
\item \textbf{RNN}: We used one \texttt{SimpleRNN} layer with $64$ neurons followed by a \texttt{Dropout} layer. The \texttt{SimpleRNN} layer uses \texttt{RELU} as the activation function and $L_1$ and $L_2$ as the regularization factor.
\item \textbf{CNN}: We stacked three \texttt{CNN1D} layers with $64$, $32$, and $16$ filters, followed by \texttt{MaxPooling1D} layer with pool size $=2$. Each \texttt{CNN1D} layer uses \texttt{RELU} as the activation function, ``same'' padding mode sand 2 as the kernel size.
\item \textbf{GNN}: We consider each time series as one node and the connection as the edge. For the variables in Figure \ref{fig:domain}, we assign ``1'' to the edge between two connected nodes, otherwise, ``0''. For the architecture, we stacked two graph convolutional networks (GCNs) with $32$ and $16$ neurons. In parallel, a \texttt{LSTM} layer with $32$ units is used to learn the temporal dynamics from the input. We then concatenate the embedding from \texttt{GCN} and \texttt{LSTM} and output the outcomes.
\item \textbf{TCN}: It is a temporal dilated convolutional network with an exponentially large receptive field. We used $64$ as the number of filters, $2$ as the kernel size, $[1, 2, 4, 8, 16, 32]$ as dilations, and 0.1 as the dropout rate. Subsequently, we added two \texttt{Dense} layers to compute the output. \texttt{LayerNormalization} is employed after each layer.
\item \textbf{RCNN}: A \texttt{SimpleRNN} with $64$ neurons, \texttt{RELU} as the activation function and $L_1$ and $L_2$ as the regularization factor is used first. 
We then used a \texttt{CNN1D} layer with $32$ filters to deal with the embedding above.
The \texttt{CNN1D} layer has 2 as kernel size, ``same'' padding mode, and \texttt{RELU} as the activation function. 
\texttt{MaxPooling1D} layer with pool size $=2$ is added after \texttt{CNN1D} layer. 
\item \textbf{Transformer}: The standard Transformer encoder \cite{vaswani2017attention} is used. 
In our case, we only used 1 \texttt{Encoder} with 1 \texttt{head}. 
We set \texttt{head size}=128, \texttt{embedding dimension}=64, \texttt{Dense} layer with 32 neurons. \texttt{Dropout} is used after each layer above. 
\item \textbf{GTN}: We combine the components of GNNs, LSTMs, Transformer, CNNs, and attention, as described in Figure \ref{fig:graphtransformer}. We have $32$ and $16$ for GNN's channels, $16$ for LSTM units, $96$ for CNN filters, and one Transformer encoder with $3$ heads (\texttt{head size}=192, \texttt{embedding dimension}=96). \\
\end{itemize}

\begin{figure}[ht]
\centering
    \includegraphics[width=0.32\columnwidth]{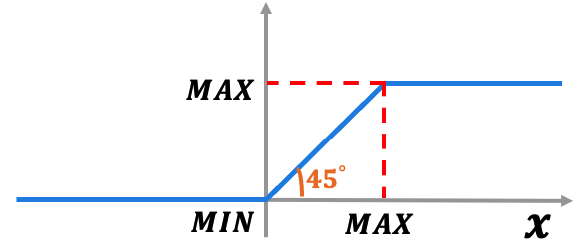} 
\caption{Constraint for the values gate and pump openings.}
\label{fig:hard_constraint}
\end{figure}

\paragraph{Note.} 
When we used the above models for \texttt{Flood Evaluator} and \texttt{Flood Manager}, the input and output need to be changed accordingly (see Figures \ref{fig:flood_evaluator} and \ref{fig:flood_manager}). To align with real-world scenarios, we enforce stringent constraints that restrict the output of \texttt{Manager} (values for gate and pump openings) to fall within the range of $[MIN, MAX]$ by a modified \texttt{RELU} activation function in Eq. (\ref{eq:hard_constraint}). Figure \ref{fig:hard_constraint} provides the corresponding visualization.
\begin{equation}
\label{eq:hard_constraint}
    x = min\{max\{x, MIN\}, MAX\},
\end{equation}
where $x$ is the opening value of gates and pumps.

\subsection{Training hyper-parameters}
In the following table, we show some hyper-parameters used in our experiments: learning rate, batch size, epoch, decay rate, decay step, patience for early stopping, dropout rate, and regularization factors $L_1$ and $L_2$.
\begin{table}[ht]
\centering
\caption{Best hyperparameters of each DL model for \texttt{Flood Evaluator}.}
    \begin{tabular}{cccccccccc}
        \toprule
        \textbf{Methods}    & \textbf{LR}   & \textbf{Batch Size} & \textbf{Epoch}   & \textbf{Decay Rate}  & \textbf{Decay Step}   & \textbf{Patience} & \textbf{Dropout Rate} & \textbf{$L_1$}   & \textbf{$L_2$}  \\
        \midrule\midrule
        MLP          & 1e-3     & 512    & 3000    & 0.95   & 10,000    & 500  & 0.1  & 1e-5  & 1e-5 \\  
        RNN          & 1e-4     & 512    & 3000    & 0.95   & 10,000    & 500  & 0.2  & 1e-5  & 1e-5 \\   
        CNN          & 1e-4     & 512    & 3000    & 0.95   & 10,000    & 500  & 0.1  & 0.0  & 0.0 \\ 
        GCN          & 5e-4     & 512    & 3000    & 0.95   & 10,000    & 500  & 0.0  & 0.0  & 0.0 \\ 
        TCN          & 5e-4     & 512    & 3000    & 0.95   & 10,000    & 500  & 0.1  & 0.0  & 0.0 \\  
        RCNN         & 1e-4     & 512    & 3000    & 0.95   & 10,000    & 500  & 0.2  & 0.0  & 1e-5\\
        Transformer  & 5e-4     & 512    & 3000    & 0.95   & 10,000    & 500  & 0.3  & 0.0  & 0.0 \\ 
        \midrule
        GTN (ours) & 5e-4  & 512    & 3000    & 0.95   & 10,000    & 500  & 0.5  & 1e-5  & 1e-5 \\  
        \bottomrule
    \end{tabular}
\label{tab:hyper_evaluator}
\end{table}
\begin{table*}[ht]
\centering
\caption{Best hyperparameters of each DL model for \texttt{Flood Manager}.}
    \begin{tabular}{cccccccccc}
        \toprule
        \textbf{Methods}   & \textbf{LR}  & \textbf{Batch Size}  & \textbf{Epoch}  & \textbf{Decay Rate}  & \textbf{Decay Step}  & \textbf{Dropout Rate}   & \textbf{Patience}    & \textbf{$L_{1}$}  & \textbf{$L_{2}$}  \\
        \midrule \midrule     
        MLP               & 1e-3      & 512   & 700   & 0.95    & 10,000    & 0.0   & 100    & 1e-5   & 1e-5     \\  
        RNN               & 1e-3      & 512   & 700   & 0.95    & 10,000    & 0.2   & 100    & 1e-5   & 1e-5   \\   
        CNN               & 3e-3      & 512   & 700   & 0.95    & 10,000    & 0.0   & 100    & 0.0    & 0.0    \\ 
        GNN               & 5e-4      & 512   & 700   & 0.9     & 10,000    & 0.0   & 100    & 0.0    & 0.0     \\ 
        TCN               & 1e-3      & 512   & 700   & 0.9     & 10,000    & 0.0   & 100    & 0.0    & 0.0     \\  
        RCNN              & 1e-3      & 512   & 700   & 0.9     & 10,000    & 0.0   & 100    & 0.0    & 1e-5   \\
        Transformer       & 5e-4      & 512   & 700   & 0.9     & 10,000    & 0.5   & 100    & 0.0    & 0.0     \\ 
        GTN (ours)   & 3e-3      & 512   & 700   & 0.95    & 10,000    & 0.5   & 100    & 1e-5   & 1e-5     \\
        \bottomrule
    \end{tabular}
\label{tab:hyper_manager}
\end{table*}

\section{More experimental results}
\label{sec:more_exp}
\subsection{Flood prediction at all locations}
\label{sec:more_exp_flood_pred}
In Table \ref{tab:flood_prediction_s1}, we provided the comparison of the performance of models for the \texttt{Flood Evaluator} at time point $t+1$, but at one location (S1).
In Table \ref{tab:flood_prediction}, we provide the comparison of the performance of different models for the \texttt{Flood Evaluator} on the test set at time $t+1$, but for all locations. 
Our GTN model still provides the best performance in flood prediction.
\begin{table*}[ht]
\centering
\small
\caption{Comparison of different models for the \texttt{Flood Evaluator} on the test set (at time t+1 for all stations). ``Over Timesteps'' (and ``Under Timesteps'') represent the number of time steps during which the water levels exceed the upper threshold (subceed the lower threshold, resp.). Similarly, ``Over Area'' (and ``Under Area'') refer to the area between the water level curve and the upper threshold (lower threshold, resp.), as shown in Figure \ref{fig:loss} in the manuscript.}
    \begin{tabular}{l|cc|cccc}
    \toprule
    \textbf{Methods}    & \textbf{MAE (ft)}   & \textbf{RMSE (ft)} & \textbf{Over Timesteps}   & \textbf{Over Area}  & \textbf{Under Timesteps}   & \textbf{Under Area} \\
    \midrule\midrule
    Ground-truth      & -            & -        & \textcolor{blue}{427}      & \textcolor{blue}{68.61}    & \textcolor{blue}{5,116}    & \textcolor{blue}{1465.10}  \\ 
    \midrule
    HEC-RAS           & 0.185        & 0.238    & 334      & 50.85    & 4,469    & 1274.26  \\ 
    \midrule
    MLP               & 0.060        & 0.080    & 900      & 245.75   & 5,493    & 1536.22  \\  
    RNN               & 0.053        & 0.071    & 377      & 57.32    & 4,813    & 1308.28  \\   
    CNN               & 0.075        & 0.105    & 342      & 53.69    & \textcolor{red}{5,182}    & 1394.05  \\ 
    GCN               & 0.052        & 0.069    & 493      & 81.28    & 5,170    & 1454.78  \\ 
    TCN               & 0.063        & 0.097    & 335      & 48.37    & 5,479    & \textcolor{red}{1470.41}  \\  
    RCNN              & 0.121        & 0.138    & 288      & 45.10    & 5,631    & 1555.63  \\
    Transformer       & 0.049        & 0.065    & 515      & 85.49    & 5,046    & 1372.50  \\ 
    \midrule
    GTN (ours)   & \textcolor{orange}{0.047}        & \textcolor{orange}{0.064}    & \textcolor{red}{439}      & \textcolor{red}{76.01}    & 5,859     & 1689.45  \\  
    \bottomrule
    \end{tabular}
\label{tab:flood_prediction}
\end{table*}

\subsection{Flood mitigation at all locations}
\label{sec:more_exp_flood_miti}
In Table \ref{tab:flood_mitigate_s1}, we provided the comparison of the performance of different models for the \texttt{Flood Manager} at time point $t+1$, but at one location (S1). We provide the comparison results of different models for the \texttt{Flood Manager} on the test set (at time t+1 for all locations) in Table \ref{tab:flood_mitigate}.
All DL-based methods show better results than ``rule-based'' and GA-based approaches for flood mitigation. Our GTN model provides the best performance in flood mitigation and comparable performance in avoiding water wastage.
\begin{table*}[ht]
\centering
\small
\caption{Comparison of different methods for the \texttt{Flood Manager} on the test set (at time t+1 for all stations). The $^{*}$ denotes that the GA method was used for \texttt{Flood Manager} with a physics-based (HEC-RAS) as an \texttt{Evaluator}. $-$ denotes too long to get results. The $^{\dag}$ denotes that the GA method was used for \texttt{Flood Manager} with the DL-based GTN as the \texttt{Evaluator}. All other rows are DL-based flood managers with a DL-based GTN as the \texttt{Evaluator}. We mark the best results in bold.} 
\label{tab:flood_mitigate}
    \begin{tabular}{l|c|cccc}
    \toprule
    \multirow{1}{*}{\textbf{Method}}
        & \textbf{Manager}        & \textbf{Over Timesteps}    & \textbf{Over Area}   & \textbf{Under Timesteps}    & \textbf{Under Area}  \\
    \midrule \midrule 
    \multirow{1}{*}{Rule-based}
        & -                 & 427         & 68.61    & 5,116      & 1465.10  \\  
    \midrule
    \multirow{2}{*}{GA-based}    
        & Genetic Algorithm$^{*}$     & -          & -         & -         & -  \\
        & Genetic Algorithm$^{\dag}$  & 343         & 67.50    & 1,817      & 416 \\
    \midrule
    \multirow{8}{*}{DL-based}       
        & MLP          & 435         & 67.28    & 3,511      & 818.06  \\  
        & RNN          & 211         & 25.28    & \textbf{386}  & \textbf{58.39}  \\   
        & CNN          & 372         & 54.72    & 3,969       & 1,044.71  \\ 
        & GCN          & 181         & 25.06    & 1,109       & 207.48  \\ 
        & TCN          & 224         & 25.61    & 562         & 92.16  \\  
        & RCNN         & 170         & 20.01    & 542         & 88.72  \\
        & Transformer  & 374         & 54.31    & 4,378       & 1,142.98  \\ 
        & GTN (Ours)   & \textbf{114}  & \textbf{12.09}    & 617         & 98.42  \\
    \bottomrule
    \end{tabular}
\end{table*}

\subsection{Flood mitigation with one event}
\label{sec:flood_mitigate_event_s1}
In Figure \ref{fig:visualize_flood_mitigation_s1}, we visualize the results of different models for the \texttt{Flood Manager} on a short event spanning 18 hours from September 3rd (09:00) to September 4th (03:00) in 2019 for one location of interest. Below are the corresponding experiment results. Our GTN shows the best performance in flood mitigation.
\begin{table*}[ht!]
\centering
\small
\caption{Comparison of different \texttt{Managers} for flood Mitigation (at time t+1 for one event S1). The best results are in bold.}
    \begin{tabular}{lccccc}
        \toprule
        \multirow{1}{*}{\textbf{Method}}
                & \textbf{Manager}   & \textbf{Over Timesteps}    & \textbf{Over Area}  & \textbf{Under Timesteps}    & \textbf{Under Area}\\
        \midrule \midrule     
        \multirow{1}{*}{Rule-based} & -     & 6      & 0.866       & 0      & 0 \\
        \midrule
        \multirow{2}{*}{GA-Based}   & Genetic Algorithm$^{*}$ (GA)   & 4    & 0.351    & 0    & 0  \\
                                    & Genetic Algorithm$^{\dag}$ (GA) & 6   & 0.764    & 0    & 0 \\
        \midrule
        \multirow{8}{*}{DL-Based}   & MLP          & 6      & 0.614      & 0      & 0 \\  
                                    & RNN          & \textbf{1}      & 0.074      & 0      & 0 \\   
                                    & CNN          & 6      & 0.592      & 0      & 0 \\ 
                                    & GNN          & 2      & 0.062      & 0      & 0 \\ 
                                    & TCN          & \textbf{1}      & 0.046      & 0      & 0 \\  
                                    & RCNN         & \textbf{1}      & 0.045      & 0      & 0 \\
                                    & Transformer  & 6      & 0.614      & 0      & 0 \\ 
                                    & GTN (Ours)     & \textbf{1}      & \textbf{0.022}      & 0      & 0 \\
        \bottomrule
    \end{tabular}
\label{tab:flood_mitigate_event_s1}
\end{table*}

\newpage
\section{Visualization}
\label{sec:vis_appendix}
\begin{figure}[ht]
\centering
    \includegraphics[width=0.95\columnwidth]{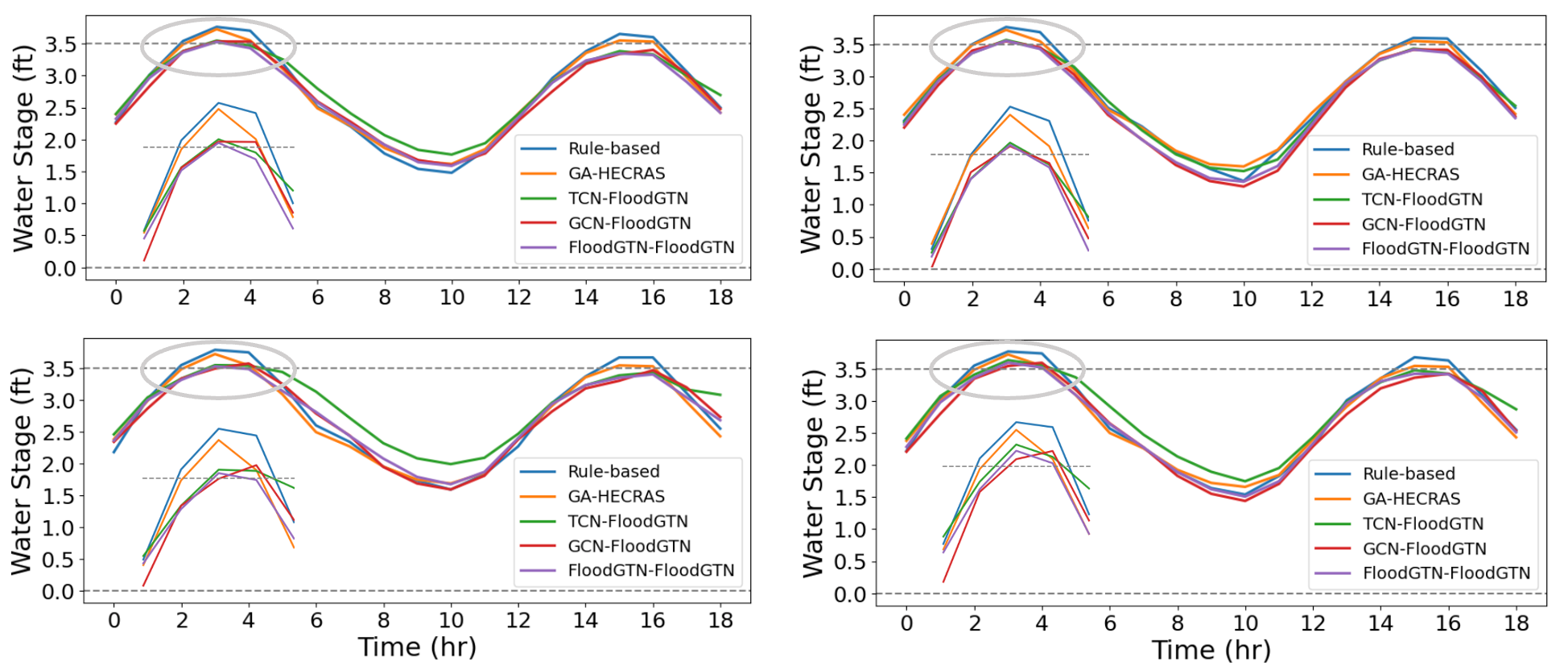} 
\caption{Visualization for flood mitigation at \textbf{all} locations. Two gray dashed lines represent the upper (3.5 ft) and lower threshold (0.0 ft).}
\label{fig:visualize_flood_mitigation_all_locations}
\end{figure}

Figure \ref{fig:visualize_flood_mitigation_all_locations} presents the results for \textbf{all} locations compared to Figure \ref{fig:visualize_flood_mitigation_s1} for one location of interest. Table \ref{tab:flood_mitigate_event} shows the corresponding performance measures. Our GTN shows the best performance in flood mitigation.

\begin{table}[ht]
\centering
\caption{Comparison of different methods of \texttt{Flood Manager} for flood Mitigation (at time t+1 for one event at \textbf{all} locations). We mark the best results in bold.}
\begin{tabular}{lccccc}
\toprule
\multirow{1}{*}{\textbf{Method}}
        & \textbf{Manager}    & \textbf{Over Timesteps}    & \textbf{Over Area}   & \textbf{Under Timesteps}    & \textbf{Under Area}  \\
\midrule \midrule     
\multirow{1}{*}{Rule-based}
        & -            & 23        & 3.62     & 0      & 0  \\
\midrule
\multirow{2}{*}{GA-Based}   & Genetic Algorithm$^{*}$ (GA)            & 16        & 1.39     & 0      & 0  \\
                            & Genetic Algorithm$^{\dag}$ (GA)          & 23        & 3.75     & 0      & 0  \\
\midrule
\multirow{8}{*}{DL-Based}   & MLP          & 23        & 4.18     & 0      & 0  \\ 
                            & RNN          & 8         & 0.63     & 0      & 0  \\   
                            & CNN          & 21        & 2.92     & 0      & 0  \\ 
                            & GCN          & 6         & 0.32     & 0      & 0  \\ 
                            & TCN          & 6         & 0.39     & 0      & 0  \\  
                            & RCNN         & 6         & 0.39     & 0      & 0  \\
                            & Transformer  & 22        & 3.00     & 0      & 0  \\ 
                            & GTN$^{\ddag}$ (Ours)     & \textbf{5}         & \textbf{0.22}     & 0      & 0  \\
\bottomrule
\end{tabular}
\label{tab:flood_mitigate_event}
\end{table}

\end{document}